\pdfoutput=1

\documentclass[11pt]{article}

\usepackage[preprint]{acl}

\usepackage{times}
\usepackage{latexsym}

\usepackage[T1]{fontenc}

\usepackage[utf8]{inputenc}

\usepackage{microtype}

\usepackage{inconsolata}

\usepackage{graphicx}

\usepackage{hyperref}       
\usepackage{url}            
\usepackage{booktabs}       
\usepackage{amsfonts}       
\usepackage{nicefrac}       
\usepackage{xcolor}         
\usepackage{multirow}
\usepackage{comment}
\usepackage{wrapfig}
\usepackage{tablefootnote}
\usepackage{caption}
\usepackage{subcaption}
\usepackage{colortbl}
\usepackage{amsmath}
\usepackage[normalem]{ulem}
\useunder{\uline}{\ul}{}

\title{Intent Detection in the Age of LLMs}

\author{Gaurav Arora \\
  Amazon \\
  \texttt{gaurvar@amazon.com} \\\And
  Shreya Jain \\
  IIT Jammu\thanks{Contributed to this work during her internship at Amazon} \\
  \texttt{2020uee0135@iitjammu.ac.in} \\\And
  Srujana Merugu \\
  Amazon \\
  \texttt{smerugu@amazon.com} \\}

\begin{document}
\maketitle
\begin{abstract}

Intent detection is a critical component of task-oriented dialogue systems (TODS) which enables the identification of suitable actions to address user utterances at each dialog turn. Traditional approaches relied on computationally efficient supervised sentence transformer encoder models, which require substantial training data and struggle with out-of-scope (OOS) detection. The emergence of generative large language models (LLMs) with intrinsic world knowledge presents new opportunities to address these challenges.
In this work, we adapt 7 SOTA LLMs using adaptive in-context learning and chain-of-thought prompting for intent detection, and compare their performance with contrastively fine-tuned sentence transformer (SetFit) models to highlight  prediction quality and latency tradeoff. We propose a hybrid system using uncertainty based routing strategy to combine the two approaches that along with negative data augmentation results in achieving the best of both worlds ( i.e. within 2\% of native LLM accuracy with 50\% less latency). 
To better understand LLM OOS detection capabilities, we perform controlled experiments revealing that this capability is significantly influenced by the scope of intent labels and the size of the label space. We also introduce a two-step approach utilizing internal LLM representations, demonstrating empirical gains in OOS detection accuracy and F1-score by >5\% for the Mistral-7B model.

\end{abstract}

\section{Introduction}
\label{sec:intro}

Task oriented dialogue systems (TODS) have gained significant traction and investment from industry because of their efficiency, accessibility and 24x7 availability to serve customers. Automation through TODS is expected to save billions of dollars in labor costs by 2026 \cite{gartner_article}. 

Intent Detection is a vital part of natural language understanding (NLU) layer of TODS. Traditionally, intent detection has been used to understand and map the user query to a bot action (e.g., respond with a static answer, execute a pre-configured flow etc) \cite{dialogflow_intent, lex_intent}. With increasing use of LLMs such as ChatGPT \cite{chatgpt_website}, Claude \cite{claude_website}, Mistral \cite{mistral_website}, Llama \cite{llama_website}  as retrieval augmented generators to generate answers to user queries in TODS, intent detection is being used to identify the right knowledge sources, APIs, and tools to call for retrieval augmented generation. This ensures efficient utilization of tools, APIs and various other knowledge sources.

 \begin{figure}
    \centering
  \includegraphics[width=\columnwidth]{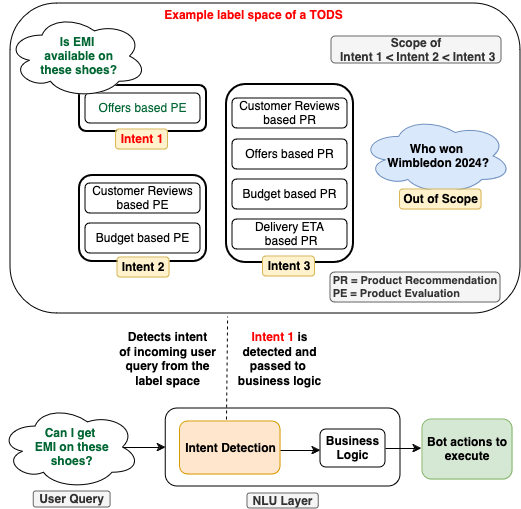}
  \caption{Example of broad/specific intent scopes and OOS queries which Intent Detection systems deal with in a typical TODS.}
  \label{fig:intro_example}
\end{figure}

\begin{figure*}
\centering
\includegraphics[width=\textwidth]{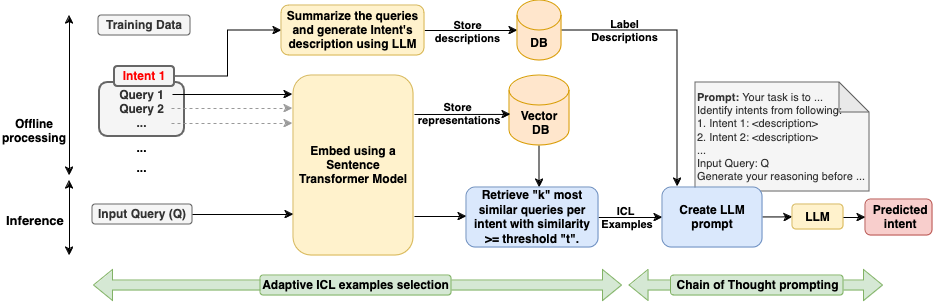}
\caption{Methodology for adaptive ICL and CoT based intent detection using LLMs.}
\label{fig:rag_intent}
\end{figure*}

 An intent detection system of a conversational AI service is expected to handle intents anywhere in the spectrum of \underline{very-broad to very-specific} scopes\footnote{By "scope of intent" we mean semantic space of all natural language utterances which can fall in that intent.} depending upon actionability of intents and bot usecases as shown in Fig \ref{fig:intro_example}. They are also expected to accurately \underline{reject out-of-scope} (OOS) queries\footnote{Out-of-scope (OOS) queries are the ones which do not fall into any of the system’s supported intents \cite{larson-etal-2019-evaluation}.} without having access to any training data for such queries as universe of OOS queries for any TODS is infinitely large. Since for a typical conversational AI service, data for intent detection training comes from bot developers who are not experts in ML, intent detection systems have to also deal with \underline{imbalanced training datasets}. Additionally, these systems are expected to work with very \underline{few utterances per intent}.

Traditionally, intent detection systems have been built using supervised classification or similarity based models \cite{pretraining_intent_zhang, rnn_attn_intent_detection_liu, casanueva-etal-2020-efficient}. LLMs, due to their few-shot learning capabilities, world knowledge and impressive performance across multiple NLP tasks \cite{qin2023chatgpt, zhao2023survey}, have the potential to improve intent detection systems in TODS. In this work, we explore how LLMs can be best leveraged for the task of intent detection and assess their ability to handle OOS queries and varying scope of intents. \\
\noindent\textbf{Contributions.}
 \noindent 1. We employ generative LLMs using adaptive in-context learning (ICL) and chain of thought (CoT) prompting for the task of intent detection and compare them against contrastively fine-tuned sentence transformer (SetFit) models, highlighting performance/latency trade-offs. We evaluate 7 SOTA LLMs from Claude and Mistral families on 3 open-source and 3 internal real world 
 datasets.
 
\noindent 2. We propose a hybrid system that combines SetFit and LLM by conditionally routing queries to LLM based on SetFit's predictive uncertainty determined using Monte Carlo Dropout. We also propose a  negative data augmentation technique that improves SetFit's performance by $>$5\% across datasets. The resulting system achieves performance within $\sim$2\% of native LLM performance with $\sim$50\% less latency than native LLM.

\noindent 3. We study the behavior of adaptive ICL based intent detection through controlled experiments and show that LLM's OOS detection capability significantly depends upon the scope of intent labels (class design) and the number of labels.

\noindent 4. We also propose a novel two step methodology utilizing internal LLM representations to help improve LLM's OOS detection capabilities and show empirical gains in OOS detection accuracy and 
F1-score by >5\% across datasets for Mistral-7B.

We intend to also share the three internal datasets after necessary approvals as a community resource and to ensure reproducibility.

\section{Related Work}
\label{sec:related_work}
\begin{table*}[]
\centering
\resizebox{0.98\textwidth}{!}{%
\begin{tabular}{lcccccccc}
\hline
\textbf{}                  & \textbf{\begin{tabular}[c]{@{}c@{}}SOF\\ Mattress\end{tabular}} & \textbf{Curekart} & \textbf{\begin{tabular}[c]{@{}c@{}}Power\\ Play11\end{tabular}} & \textbf{ALC}   & \textbf{ADP}   & \textbf{OADP}  & \textbf{\begin{tabular}[c]{@{}c@{}}Avg\\  Score\end{tabular}} & \textbf{\begin{tabular}[c]{@{}c@{}}Avg p50\\  Latency\end{tabular}} \\ \hline
\textbf{Claude v1 Instant} & 0.613                                                           & 0.528             & 0.295                                                           & 0.840          & 0.687          & 0.630          & 0.599                                                         & 2.297                                                                    \\
\textbf{Claude v2}         & 0.763                                                           & 0.773             & {\ul 0.665}                                                     & 0.891          & 0.703          & {\ul 0.630}    & \textbf{0.737}                                                & 11.795                                                                   \\
\textbf{Claude v3 Haiku}   & \textbf{0.815}                                                  & {\ul 0.775}       & 0.646                                                           & 0.849          & {\ul 0.715}    & 0.619          & {\ul 0.736}                                                   & 1.697                                                                    \\
\textbf{Claude v3 Sonnet}  & 0.739                                                           & 0.647             & 0.566                                                           & {\ul 0.895}    & \textbf{0.765} & \textbf{0.653} & 0.711                                                         & 4.592                                                                    \\
\textbf{Mistral 7B}        & 0.699                                                           & 0.615             & 0.384                                                           & 0.804          & 0.624          & 0.453          & 0.597                                                         & 1.624                                                                    \\
\textbf{Mixtral 8x7B}      & 0.694                                                           & 0.614             & 0.434                                                           & 0.824          & 0.653          & 0.587          & 0.634                                                         & 1.992                                                                    \\
\textbf{Mistral Large}     & {\ul 0.767}                                                     & \textbf{0.779}    & \textbf{0.668}                                                  & \textbf{0.907} & 0.688          & 0.601          & 0.735                                                         & 3.565                                                                    \\
\textbf{SetFit (Baseline)} & 0.632                                                           & 0.511             & 0.612                                                           & 0.769          & 0.617          & 0.462          & 0.600                                                         & 0.030                                                                    \\
\textbf{SetFit + Neg Aug}  & 0.672                                                           & 0.709             & 0.639                                                           & 0.848          & 0.625          & 0.459          & 0.658                                                         & 0.030                                                                    \\ \hline
\end{tabular}%
}
\caption{Comparison of F1 Score of various SOTA LLMs with fine tuned sentence transformer models across AID3 and HINT3 datasets}
\label{tab:benchmark_res}
\end{table*}

\textbf{Evaluation of LLMs.} LLMs like ChatGPT \cite{chatgpt_website}, GPT-4 \cite{openai2024gpt4}, Claude \cite{claude_website}, Mistral \cite{mistral_website}, Llama \cite{llama_website} have shown impressive performance on multiple NLP tasks and benchmarks \cite{zhao2023survey}. Supervised BERT \cite{bert_paper} based models have been widely used for intent detection but now with the advent of LLMs it is not clear what benefits they bring for intent detection in the real world. Hence in this work, we evaluate LLMs on the critical task of intent detection for TODS on real world intent detection datasets and highlight performance/latency tradeoffs by benchmarking LLMs with traditional sentence transformers. Recent work \cite{wang2024knowninvestigatingllmsperformance, liu2024goodllmsoutofdistributiondetection} majorly focused on evaluation of LLMs on datasets like CLINC150 \cite{larson-etal-2019-evaluation}, BANKING77 \cite{casanueva-etal-2020-efficient} which are: (i) not real world intent detection datasets (queries are not from deployed TODS), (ii) not multi-label (every query maps to single intent). Instead, our evaluation is on real world intent detection datasets wherein queries are from deployed TODS which have real world challenges like intents with very-broad to very-specific scopes, imbalanced training datasets with very few examples per intent and 3 out of 6 of our datasets are also multi-label which makes our evaluation more comprehensive.

\textbf{Improving OOS detection performance of LLMs.} Recent work \cite{liu2024goodllmsoutofdistributiondetection} fine-tuned LLMs to improve OOS performance which is prohibitive both from development and maintenance perspective for a typical Conversational-AI platform which needs to support hundreds of different TODS (because fine-tuning and deploying a separate instance of LLM for every TODS is prohibitively expensive which makes fine-tuning LLMs impractical). Hence, we propose an alternative approach without LLM fine-tuning which improves both OOS accuracy and overall performance by >5\% and allows use of the same instance of foundational LLM across TODS.

\textbf{Hybrid intent detection system which uses LLMs.} Unlike prior work, our focus is not just on evaluation of LLMs and/or improving OOS detection performance of LLMs, but we also focus on building a deployable intent detection system which can benefit from LLMs but does not have prohibitive cost and latency, as part of which we propose a hybrid system using uncertainty based routing strategy to combine LLMs and SetFit approaches that along with negative data augmentation results in achieving the best of both worlds ( i.e. within 2\% of native LLM accuracy with 50\% less latency).

\textbf{Better understanding of LLM’s OOS detection capabilities.} In this work we do controlled experiments to study the effect of scope of labels and size of label space. Recent work \cite{wang2024knowninvestigatingllmsperformance} also investigated the effect of the size of the label space on LLM's OOS performance and their findings are inline with our findings. However, our findings on how LLM OOS detection capabilities are influenced by the scope of intent labels are novel and would inform label space design during development of TODS.

%


\section{Leveraging LLMs for Intent Detection}
\label{sec:llm_eval}
In this section we see how LLMs can be best leveraged for intent detection and propose a hybrid system which leverages LLMs conditionally, achieving a balance between performance and cost. 

\subsection{Methodology}

\subsubsection{Fine-Tuned Sentence Transformers}

We fine tune sentence transformer (SetFit) models in two steps \cite{tunstall2022efficientfewshotlearningprompts} and use them as our baseline. In the first step, a sentence transformer model is fine-tuned on the training data in a contrastive, siamese manner on sentence pairs. In the second step, a text classification head is trained using the encoded training data generated by the fine-tuned sentence transformer from the first step.

\textbf{Negative Data Augmentation.} To help SetFit learn better decision boundaries, we augment training data by modifying keywords in sentences by (a) removing, or (b) replacing them with random strings. These modified sentences are considered OOS during training. Since these augmented OOS sentences have similar lexical pattern as in-scope training sentences, these are expected to help the model avoid latching onto any spurious patterns and help overall learning.

\subsubsection{Adaptive ICL + CoT based Intent Detection using LLMs}
\label{sec:llm_rag_methodology}

Fig \ref{fig:rag_intent} shows how we use LLMs with adaptive ICL and CoT prompting for intent detection. During offline processing, we embed all training examples using a sentence transformer model and store the embedding vectors in a DB. Additionally, we generate and store descriptions for every intent from training data using LLM. During inference, we embed the user query using the same transformer model and retrieve top-$k$ most similar queries per intent with similarity >$t$, where $t$ is retriever threshold. We construct prompt for LLM using retrieved ICL examples, stored intent descriptions and static task specific instructions.

\subsubsection{Uncertainty based Query Routing}

High compute and latency costs of LLMs make them prohibitively expensive to use in production at scale.\footnote{Mechanisms like caching can help somewhat but we skip their discussion for brevity.} Hence, we propose a hybrid system which routes incoming queries to LLMs for intent detection only if SetFit model is uncertain. We sample $M$ predictions from the SetFit model using Monte Carlo (MC) dropout \cite{mc_dropout} and use variance of the predictions as an uncertainty estimate.

\begin{table}[]
\centering
\resizebox{0.9\columnwidth}{!}{%
\begin{tabular}{ccccc}
\hline
\multirow{3}{*}{\textbf{Dataset}} & \multirow{3}{*}{\textbf{\begin{tabular}[c]{@{}c@{}}No.\\ of \\ Intents\end{tabular}}} & \multicolumn{3}{c}{\textbf{No. of Queries}}                                                                                                                  \\ \cline{3-5} 
                                  &                                                                                          & \multirow{2}{*}{\textbf{Train}} & \multicolumn{2}{c}{\textbf{Valid}}                                                                                       \\ \cline{4-5} 
                                  &                                                                                          &                                 & \textbf{\begin{tabular}[c]{@{}c@{}}In Scope\end{tabular}} & \textbf{\begin{tabular}[c]{@{}c@{}}OOS\end{tabular}} \\ \hline
\textbf{ALC}                      & 8                                                                                        & 150                             & 338                                                         & 128                                                             \\
\textbf{ADP}                      & 13                                                                                       & 683                             & 803                                                         & 91                                                              \\
\textbf{OADP}                     & 13                                                                                       & -                               & 430                                                         & 56                                                              \\ \hline
\end{tabular}%
}
\caption{Data Statistics for AID3 dataset}
\label{tab:aid3_stats}
\end{table}

\begin{table*}[]
\centering
\resizebox{0.95\textwidth}{!}{%
\begin{tabular}{lccccccc}
\hline
\textbf{}                  & \textbf{SOFMattress} & \textbf{Curekart} & \textbf{PowerPlay11} & \textbf{ALC}   & \textbf{ADP}   & \textbf{OADP}  & \textbf{Avg Score} \\ \hline
\textbf{Claude v1 Instant} & 0.229                & 0.241             & 0.122                & 0.742          & 0.143          & 0.000          & 0.246              \\
\textbf{Claude v2}         & {\ul 0.688}          & 0.701             & 0.580                & 0.945          & 0.330          & {\ul 0.232}    & 0.579              \\
\textbf{Claude v3 Haiku}   & \textbf{0.736}       & {\ul 0.716}       & 0.561                & \textbf{0.961} & {\ul 0.593}    & 0.036          & {\ul 0.601}        \\
\textbf{Claude v3 Sonnet}  & 0.479                & 0.436             & 0.402                & {\ul 0.953}    & 0.440          & 0.036          & 0.458              \\
\textbf{Mistral 7B}        & 0.465                & 0.376             & 0.205                & 0.781          & 0.154          & 0.018          & 0.333              \\
\textbf{Mixtral 8x7B}      & 0.382                & 0.391             & 0.455                & 0.914          & 0.264          & 0.036          & 0.407              \\
\textbf{Mistral Large}     & 0.646                & \textbf{0.771}    & 0.602                & 0.945          & \textbf{0.615} & \textbf{0.268} & \textbf{0.641}     \\
\textbf{SetFit (Baseline)} & 0.563                & 0.293             & \textbf{0.798}       & 0.594          & 0.022          & 0.000          & 0.378              \\
\textbf{SetFit + Neg Aug}  & 0.681                & 0.592             & {\ul 0.665}          & 0.844          & 0.154          & 0.000          & 0.489              \\ \hline
\end{tabular}%
}
\caption{Out of Scope Recall at best F1 Score of various SOTA LLMs with fine tuned sentence transformer models across AID3 and HINT3 datasets}
\label{tab:oos_recall_res}
\end{table*}

\begin{table*}[]
\centering
\resizebox{\textwidth}{!}{%
\begin{tabular}{lccccccccccccc}
\hline
\textbf{}                                                                            &                              &                                                                                   &                                                                                &                                                                                   &                                &                                &                                 & \multicolumn{2}{c}{\textbf{Avg score}}                                   &                                                                                   & \multicolumn{2}{c}{\textbf{Delta Avg score}}                            &                                                                                        \\ \cline{9-10} \cline{12-13}
\textbf{}                                                                            & \multirow{-2}{*}{\textbf{M}} & \multirow{-2}{*}{\textbf{\begin{tabular}[c]{@{}c@{}}SOF\\ Mattress\end{tabular}}} & \multirow{-2}{*}{\textbf{\begin{tabular}[c]{@{}c@{}}Cure\\ Kart\end{tabular}}} & \multirow{-2}{*}{\textbf{\begin{tabular}[c]{@{}c@{}}Power\\ Play11\end{tabular}}} & \multirow{-2}{*}{\textbf{ALC}} & \multirow{-2}{*}{\textbf{ADP}} & \multirow{-2}{*}{\textbf{OADP}} & \textbf{} & \textbf{\begin{tabular}[c]{@{}c@{}}w/o \\ OADP\end{tabular}} & \multirow{-2}{*}{\textbf{\begin{tabular}[c]{@{}c@{}}Avg\\  latency\end{tabular}}} & \textbf{} & \textbf{\begin{tabular}[c]{@{}c@{}}w/o\\ OADP\end{tabular}} & \multirow{-2}{*}{\textbf{\begin{tabular}[c]{@{}c@{}}Latency\\  fraction\end{tabular}}} \\ \hline
\textbf{SNA}                                                                         & \textbf{-}                   & 0.672                                                                             & 0.709                                                                          & 0.639                                                                             & 0.848                          & 0.625                          & 0.459                           & 0.658     & 0.698                                                        & 0.030                                                                             & -0.078    & -0.061                                                      & 0.013                                                                                  \\ \hline
\rowcolor[HTML]{EFEFEF} 
\textbf{v3 Haiku}                                                                    & \textbf{-}                   & 0.815                                                                             & 0.775                                                                          & 0.646                                                                             & 0.849                          & 0.715                          & 0.619                           & 0.736     & 0.760                                                        & 2.345                                                                             & 0.000     & 0.000                                                       & 1.000                                                                                  \\
                                                                                     & 5                            & 0.719                                                                             & 0.734                                                                          & 0.654                                                                             & 0.849                          & 0.653                          & 0.473                           & 0.680     & 0.722                                                        & 0.748                                                                             & -0.056    & -0.038                                                      & 0.319                                                                                  \\
                                                                                     & 10                           & 0.740                                                                             & 0.747                                                                          & 0.671                                                                             & 0.863                          & 0.666                          & 0.489                           & 0.696     & 0.737                                                        & 1.005                                                                             & -0.040    & -0.022                                                      & 0.429                                                                                  \\
\multirow{-3}{*}{\textbf{\begin{tabular}[c]{@{}l@{}}SNA +\\ v3 Haiku\end{tabular}}}  & 20                           & 0.730                                                                             & 0.756                                                                          & 0.690                                                                             & 0.855                          & 0.668                          & 0.485                           & 0.697     & 0.740                                                        & 1.287                                                                             & -0.039    & -0.020                                                      & 0.549                                                                                  \\ \hline
\rowcolor[HTML]{EFEFEF} 
\textbf{Mistral-L}                                                                   & -                            & 0.767                                                                             & 0.779                                                                          & 0.668                                                                             & 0.907                          & 0.688                          & 0.601                           & 0.735     & 0.762                                                        & 3.867                                                                             & 0.000     & 0.000                                                       & 1.000                                                                                  \\
                                                                                     & 5                            & 0.712                                                                             & 0.739                                                                          & 0.648                                                                             & 0.872                          & 0.651                          & 0.481                           & 0.684     & 0.724                                                        & 1.063                                                                             & -0.051    & -0.037                                                      & 0.275                                                                                  \\
                                                                                     & 10                           & 0.726                                                                             & 0.747                                                                          & 0.668                                                                             & 0.879                          & 0.662                          & 0.497                           & 0.696     & 0.736                                                        & 1.453                                                                             & -0.038    & -0.025                                                      & 0.376                                                                                  \\
\multirow{-3}{*}{\textbf{\begin{tabular}[c]{@{}l@{}}SNA +\\ Mistral-L\end{tabular}}} & 20                           & 0.719                                                                             & 0.761                                                                          & 0.692                                                                             & 0.872                          & 0.664                          & 0.498                           & 0.701     & 0.742                                                        & 1.657                                                                             & -0.034    & -0.020                                                      & 0.428                                                                                  \\ \hline
\end{tabular}%
}
\caption{Table showing F1 score of two best LLMs (Claude v3 Haiku and Mistral Large) and SetFit + Neg Aug (SNA) hybrid system with varying number of samples (M) from MC dropout.}
\label{tab:mcd_ensemble}
\end{table*}

\subsection{Datasets}

We use SOFMattress, Curekart and Powerplay11 datasets from HINT3 \cite{hint3}. 
We also use \textbf{AID3}\footnote{The splits of all three datasets in AID3 were prepared specifically for experiments done as part of this work and performance on them does not reflect our production system's performance.}, a collection of three internal multi-label datasets shown in Table \ref{tab:aid3_stats} - ALC, ADP and OADP, each containing diverse set of PII redacted  in-scope and OOS real world queries from shopping domain. Both ALC and ADP contain queries from deployed shopping assistant, whereas OADP contains queries from single turn QnA forum. We use OADP to test out of distribution generalization while using ADP train set. See Appendix \ref{appendix:aid3_dataset} for more details on AID3. Label space size across HINT3 and AID3 datasets varies from 8 till 59 and all these datasets are real world intent detection datasets from deployed TODS which mimic real world scenarios and production challenges like handling intents with very-broad to very-specific scopes, imbalanced training datasets with very few examples per intent. By evaluating on HINT3 and AID3 datasets we include scenarios where there are large number of intents (59 being the maximum label space size) and also include multi-label scenarios (3 out of 6 of our datasets are also multi-label), which makes our evaluation more comprehensive.

\subsection{Experiment Setup}

\textbf{SetFit.} We use MPNet \cite{mpnet_base_v2, song2020mpnet} as the backbone and use linear layer with sigmoid as differentiable head. We do hyperparameter search over search space given in Table \ref{tab:setfit_hyperparam} using Optuna \cite{akiba2019optuna} and report best valid set results across all datasets. For MC Sampling, we use 0.1 dropout across hidden and attention layers in the backbone.

\noindent\textbf{LLMs.} We use BGE sentence transformer \cite{bge_retriever} as the retriever and do grid search over $k$ and $t$ with search space specified in Table \ref{tab:llm_hyperparam} and report best valid set results. To prevent LLMs from using any spurious patterns from intent label names, especially for open source datasets, we randomly mask them to Label-xx, where xx is some random integer. We use Claude v3 Sonnet to generate label descriptions for each intent for all datasets and keep them consistent across all LLMs.

\noindent\textbf{Metrics.} We use F1-Score as the primary performance metric. Additionally, we use OOS Recall \cite{larson-etal-2019-evaluation} and OOS AUCROC to compare model's OOS detection capabilities and use in-scope accuracy to compare their in-scope performance. 

See Appendix \ref{appendix:experiment_setup} for more details on implementation and experiment setup across models.

\subsection{Results}
\label{sec:llm_eval_results}

Evaluation results from 7 SOTA LLMs across two LLM families (Claude, Mistral) are shown in Table \ref{tab:benchmark_res}. Overall Claude v2, v3 LLMs and Mistral Large have similar performance, but Claude v3 Haiku is better amongst them with respect to latency. We see that adding negative augmentation to baseline SetFit improves performance by >5\%, but still has $\sim$8\% poor predictive performance with respect to best performing LLM. SetFit is about 56 times faster than overall best LLM (v3 Haiku). Additionally, all models see lower performance for OADP as compared to ADP but SetFit has one of the largest drop in performance ($\sim$15\%) for OADP as compared to ADP. This shows lack of generalization ability of smaller SetFit models in comparison to LLMs. Table \ref{tab:oos_recall_res} shows that all models including LLMs struggle with OOS detection with poor OOS recall across datasets.

\begin{figure*}[]
\centering
\begin{subfigure}{.5\linewidth}
  \centering
  \includegraphics[width=0.9\linewidth,height=4.1cm]{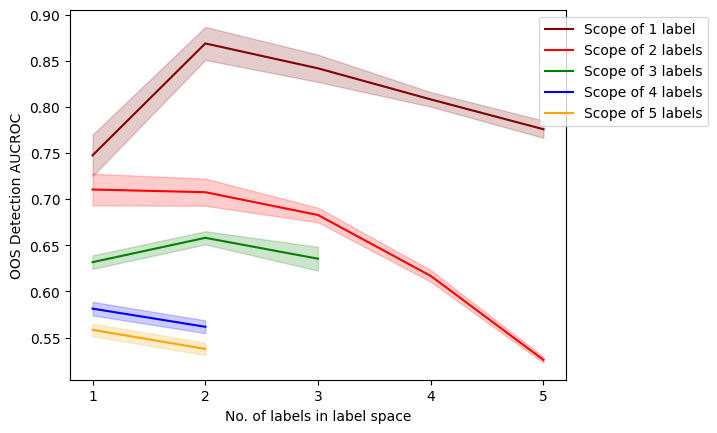}
  \caption{Claude v3 Sonnet}
  \label{fig:controlled_oos_auc_roc_claude_v3_sonnet}
\end{subfigure}%
\begin{subfigure}{.5\linewidth}
  \centering
  \includegraphics[width=0.9\linewidth,height=4.1cm]{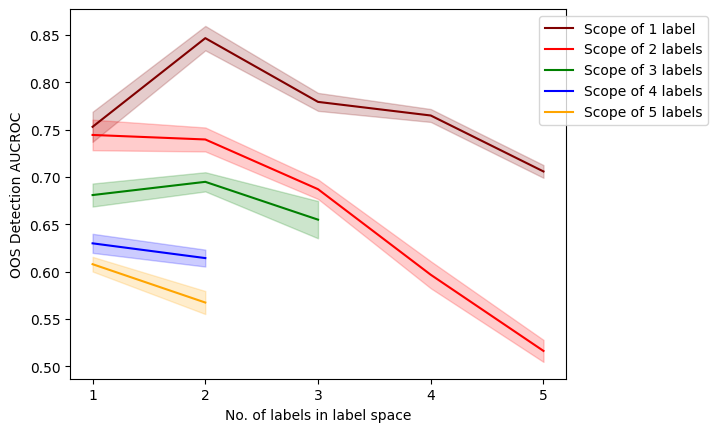}
  \caption{Claude v3 Haiku}
  \label{fig:controlled_oos_auc_roc_claude_v3_haiku}
\end{subfigure}
\begin{subfigure}{.5\linewidth}
  \centering
  \includegraphics[width=0.9\linewidth,height=4.1cm]{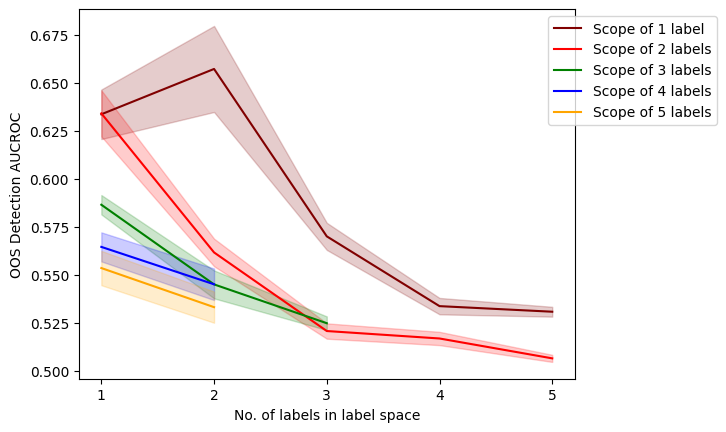}
  \caption{Mistral 7B}
  \label{fig:controlled_oos_auc_roc_mistral_7b}
\end{subfigure}%
\begin{subfigure}{.5\linewidth}
  \centering
  \includegraphics[width=0.9\linewidth,height=4.1cm]{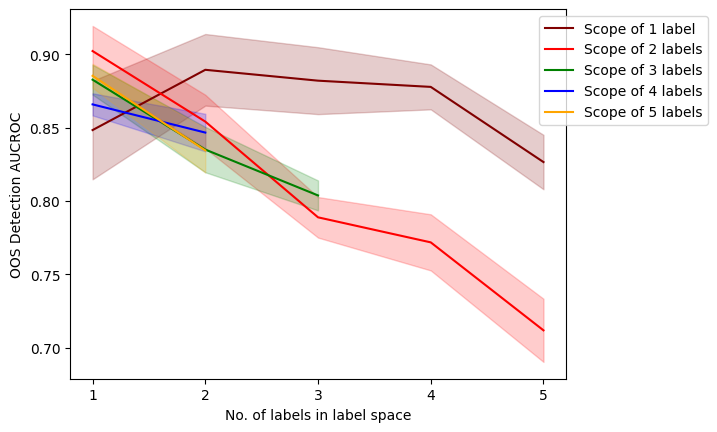}
  \caption{SetFit + Neg Aug (SNA)}
  \label{fig:controlled_oos_auc_roc_baseline}
\end{subfigure}
\caption{Change in OOS detection performance with number of labels in label space and scope of labels.}
\label{fig:controlled_oos_auc_roc}
\end{figure*}

Table \ref{tab:mcd_ensemble} shows hybrid system results for two best performing LLMs. We see that with the hybrid system we are able to bring performance gap further down to $\sim$2\% (from $\sim$6\%) for all datasets for which train and test data were from same distribution (i.e. except OADP) and down to $\sim$4\% (from $\sim$8\%) including OADP at $\sim$50\% reduced latency\footnote{Latency would reduce further if we do MC sampling in batches. See latency discussion in Appendix \ref{appendix:experiment_setup}.}. Increasing number of samples ($M$) in MC dropout does not increase performance significantly.

\section{LLMs and OOS Detection}
\label{sec:llm_oos}
Evaluation results in Sec \ref{sec:llm_eval_results} showed that LLMs struggle with OOS detection. Hence, in this section we do a controlled study to better understand behavior of LLM based intent detection with special focus on their OOS detection capabilities (Sec \ref{sec:oos_controlled_exp}) and based on the insights propose a novel methodology for OOS detection to improve LLMs performance (Sec \ref{sec:oos_two_step}).

\subsection{Analyzing LLMs OOS Detection Abilities}
\label{sec:oos_controlled_exp}

We first describe how we setup a controlled experiment to understand how varying "scope of intents" and "no. of labels" in the label space affects LLM performance, and then share our analysis results.

\noindent\textbf{Dataset.} 
We hand curate a dataset with hierarchical label space consisting of 20 leaf intents/labels and two unique parent intents as shown in Table \ref{tab:controlled_exp_dataset}. From it, we create new intents with varying scope of $S \in [1,5]$ labels by randomly combining $S$ leaf intents from the same parent, without replacement. This is realistic because in real world intent scope is driven by bot usecases and scope of APIs/systems which TODS can access.

\noindent\textbf{Experiment Setup.} 
We experiment by varying "scope of intents" by choosing intents from the newly created intents with scope of $S$ labels with $S \in [1,5]$ and experiment with varying "no. of labels" in label space by randomly picking $L$ different intents of the required scope with $L \in [1,5]$. Higher $S$ leads to intents with broader scope. We report results based on runs on 10 randomly created datasets for every experiment. See Appendix \ref{appendix:controlled_exp} for more details on the setup.

\noindent\textbf{Results and Analysis.}
Fig \ref{fig:controlled_oos_auc_roc}\footnote{Curves with scope of label > 2 are truncated because we sample and combine leaf nodes without replacement to create non-conflicting intents with bigger scope.} shows how OOS detection AUCROC for LLMs is affected with change in "scope of intents" and "no. of labels" in the label space. We see significantly more performance degradation across all LLMs in comparison to SNA model with increase in "scope of intents" and "no. of labels" in label space. This highlights greater importance of class design for LLMs and suggests that fine grained labels and smaller label spaces are better for LLM's OOS detection capabilities. From Fig \ref{fig:controlled_in_scope_acc} in Appendix \ref{appendix:controlled_exp} we see that in-scope accuracy of LLMs is relatively immune to change in "scope of intents" but degrades with increase in label space size. However, degradation in OOS detection AUCROC is worse than in-scope accuracy degradation with increase in label space size. SNA model on the other hand does show degradation in in-scope accuracy as well with both increase in "scope of intents" and "no. of labels" in label space.

\begin{figure*}
\centering
\includegraphics[width=0.98\textwidth]{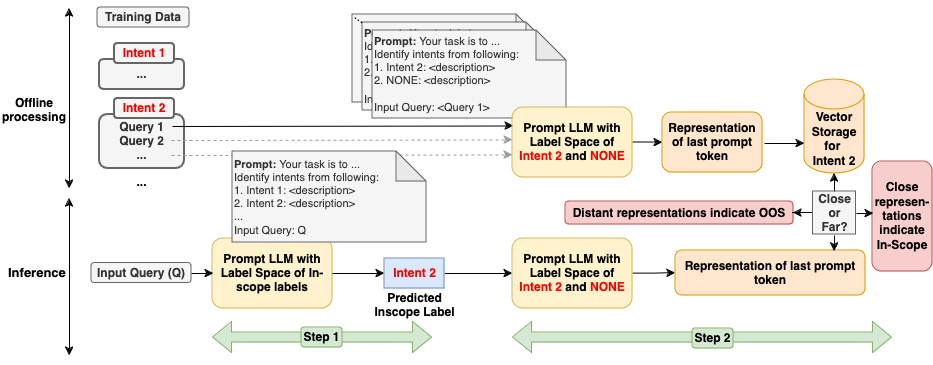}
\caption{OOS detection using LLM's internal representations}
\label{fig:two_step}
\end{figure*}

\begin{table*}[]
\centering
\resizebox{0.85\textwidth}{!}{%
\begin{tabular}{cccccc}
\hline
\textbf{}                                                                         & \textbf{}         & \textbf{Overall Accuracy}     & \textbf{F1 Score}             & \textbf{Inscope Accuracy}     & \textbf{Out of Scope Recall}  \\ \hline
                                                                                  & \textbf{Mistral-7B} & 0.705                         & 0.699                         & 0.842                         & 0.465                         \\
\multirow{-2}{*}{\textbf{\begin{tabular}[c]{@{}c@{}}SOF\\ Mattress\end{tabular}}} & \textbf{Ours}     & \cellcolor[HTML]{EBFBEA}0.748 & \cellcolor[HTML]{EBFBEA}0.751 & \cellcolor[HTML]{FFE6E5}0.767 & \cellcolor[HTML]{EBFBEA}0.715 \\ \hline
                                                                                  & \textbf{Mistral-7B} & 0.601                         & 0.615                         & 0.863                         & 0.376                         \\
\multirow{-2}{*}{\textbf{Curekart}}                                               & \textbf{Ours}     & \cellcolor[HTML]{EBFBEA}0.761 & \cellcolor[HTML]{EBFBEA}0.766 & \cellcolor[HTML]{FFE6E5}0.736 & \cellcolor[HTML]{EBFBEA}0.782 \\ \hline
                                                                                  & \textbf{Mistral-7B} & 0.357                         & 0.384                         & 0.689                         & 0.205                         \\
\multirow{-2}{*}{\textbf{\begin{tabular}[c]{@{}c@{}}Power\\ Play11\end{tabular}}} & \textbf{Ours}     & \cellcolor[HTML]{EBFBEA}0.780 & \cellcolor[HTML]{EBFBEA}0.739 & \cellcolor[HTML]{FFE6E5}0.411 & \cellcolor[HTML]{EBFBEA}0.950 \\ \hline
\end{tabular}%
}
\caption{Comparison of our two step methodology with baseline across HINT3 datasets}
\label{tab:two_step_res}
\end{table*}

\subsection{OOS Detection using LLMs Internal Representations}
\label{sec:oos_two_step}

Motivated by the insights from controlled experiment, we propose a two step methodology using LLM's internal representations to improve its performance which we describe in this section.

\subsubsection{Methodology}

Fig \ref{fig:two_step} shows our proposed methodology. During offline processing, we generate representation of each sentence in the training data by obtaining LLM decoder layer's last prompt token's representation. Then during inference, we perform following steps.

\noindent\textbf{Step 1.} Firstly, we prompt the LLM to predict one of the in-scope labels without asking it to predict out of scope by completely discarding out of scope from label space given to LLM in the prompt.

\noindent\textbf{Step 2.} Then, based on in-scope label predicted from the previous step, we generate incoming query's representation in similar way as done during offline processing using LLM's decoder layer. We then compare this representation with representations of training instances of predicted in-scope label from the first step.  

This ensures reduced label space for OOS detection but adds low latency overhead for generating representations in Step 2. But since we just need to do a forward pass for encoding the prompt, it is significantly faster than autoregressive generation.

Additionally, our proposed OOS detection methodology using LLM’s internal representations can be used to improve OOS detection performance of both fine-tuned and non-fine-tuned (base instruct tuned) LLMs. We choose to experiment and show results on non-fine-tuned LLM in Sec \ref{sec:two_step_exp} because that is a more practical scenario (as fine-tuning and deployment of a separate instance of LLM for every TODS is prohibitively expensive), but the methodology is generic enough to be used with fine-tuned LLMs as well.

\subsubsection{Experiments and Results}
\label{sec:two_step_exp}

\textbf{Setup.} We experiment with base instruct tuned Mistral-7B since its weights are open source. We use cosine similarity for comparing representations in Step 2 and take mean of scores over all training sentences of the predicted intent.

\noindent\textbf{Results.}
Table \ref{tab:two_step_res} compares results of our methodology against baseline LLM methodology discussed in Sec \ref{sec:llm_rag_methodology} for HINT3 datasets. We see >5\% improvement in performance across datasets at $\sim$300ms additional latency cost on 1 32GB V100 GPU because encoding the prompt through LLM is cheap. There is drop in in-scope performance as well but that is overcome by significant gains in OOS recall to lead to better overall performance. If needed, threshold in Step 2 of our methodology can be chosen such that drop in in-scope performance is less than an upper limit which in-turn would limit the gains in OOS performance though.


\section{Conclusion}
\label{sec:conclusion}
Various idiosyncrasies of intent detection task like varying scope of intents within a dataset, need to reject out of scope queries, imbalanced datasets and low resource regime make it a challenging task. In this work we evaluate multiple open source and closed source SOTA LLMs across multiple internal and external datasets for the task of intent detection using adaptive ICL and CoT prompting, compare them with SetFit models and discuss their performance/latency trade-offs. We build a hybrid system which routes queries to LLM when needed and achieves balance between performance and cost. We also propose a novel two step methodology which improves overall LLM performance by >5\% across datasets and share insights on how varying scope of intents and number of labels in label space affect LLM performance. We hope our work will be useful for the community to build better TODS.

\section*{Limitations}
\label{sec:limitations}
While our current work has broad applicability for the design of accurate and computationally efficient task-oriented dialog systems, there are a few limitations:

\noindent \textbf{Interactive Intent Design.} Our current work assumes that intents are specified one-time in the form of examples by human experts, which has been the norm for designing task-oriented conversational assistants. However, there is potential for leveraging LLMs for an interactive class design process. In the future, we plan to investigate the benefits of enabling domain experts to directly interact with these LLMs to interactively define and refine the scope of intents.

\noindent \textbf{Multilingual Support.} While our current empirical evaluation was primarily focused on English datasets, the SOTA LLMs we explore already provide multilingual support. To fully harness the potential of our approach, we aim to generalize our ideas to the multilingual setting and evaluate them on diverse dialog datasets across various languages.

\noindent \textbf{Alternative Hybrid Strategies.}  In the current work, we employ a cascade routing strategy that uses SetFit's uncertainty to combine the SetFit models and LLMs yielding promising results. However, there are additional hybrid strategies worth exploring. Drawing inspiration from active learning literature, we could investigate alternative utility functions, such as information gain to determine when to invoke the LLM alongside the SetFit model. We also plan to compare our approach with model distillation strategies, where the LLM is used to generate synthetic training data to enhance the SetFit models.


\section*{Ethics Statement}
\label{sec:ethics}

Our motivation for the current work is to develop computationally efficient and accurate solutions for intent detection, leveraging prior research on sentence transformers and generative language models. As the focus is on intent classification rather than generation, the typical risks associated with generative content do not directly apply. However, as with any machine learning system, there are other important considerations, such as potential biases in the training data or constituent pre-trained models, the possibility of misuse, and challenges in establishing full accountability. Since our approach incorporates generative  LLMs, any application of the proposed ideas needs to be mindful of any biases present in those models. Overall, the proposed methodological innovations are intended for benign applications and are not associated with any direct negative social impact. The datasets used in this research include public benchmarks and proprietary datasets from safe ecommerce categories, with  personally identifiable information (PII) redacted to ensure customer privacy. To enable reproducibility, we plan to share these datasets  as a community after internal approvals.

\bibliography{main}

\appendix

\section{Appendix}
\label{sec:appendix}
\subsection{AID3 Dataset}
\label{appendix:aid3_dataset}

\textbf{ALC} contains upper funnel shopping queries for 1 HCTP\footnote{High Consideration Technical Products} category while \textbf{ADP} contains lower funnel queries for 6 HCTP categories. \textbf{OADP} also contains lower funnel queries from >10 HCTP categories.

\subsection{Experiment Setup}
\label{appendix:experiment_setup}

For training SetFit models, we use SetFit library \cite{setfit_github_library} for implementation. Hyperparameter search space for SetFit model's training is given in Table \ref{tab:setfit_hyperparam}.

For \textbf{negative augmentation}, we use KeyBERT \cite{grootendorst2020keybert} for identifying keywords. For every identified keyword, random 50\% of the times we completely remove it, and remaining 50\% of the times we replace it with a randomly generated string of 5 characters. For eg: “looking for a gaming laptop” can get converted into “looking for a” or “looking for a XYCVD QSDER” or “looking for a RTYUH”. Since these augmented OOS sentences have similar lexical pattern as in-scope training sentences, these are expected to help the model avoid latching onto any spurious patterns and help overall learning, which shows up in results as well (See \ref{sec:llm_eval_results}). If $U$ is the set of randomly sampled augmentations to add to train set, then we keep $|U|$ = 0.2*$|D|$, where $|D|$ is size of train set. 

For \textbf{choosing ICL examples} for LLMs, we do grid search over ideal number of ICL examples and retriever threshold whose search space is shown in Table \ref{tab:llm_hyperparam}. We keep ordering of labels in the prompt fixed across all experiments and keep ICL examples within a label in descending order of similarity with incoming query.

For \textbf{Monte Carlo (MC) sampling} from SetFit models for hybrid system, we look at variance of the predictions as an uncertainty estimate. Specifically, let $p_i \in P \forall i \in [1,M]$ be the predicted label with maximum score from $i^{th}$ sample, where $M$ is the maximum number of samples. Then, we consider the prediction to be uncertain if number of different values of $p_i \forall i \in [1,M]$ is greater than 1 or less than $M/2$. We add upper limit of $M/2$ for stability. 

For \textbf{latency calculations of hybrid system}, we also add time for doing multiple forward passes sequentially through SetFit in MC sampling procedure keeping memory needs constant. Since maximum $M =20$ in our experiments, if we consider that sampling can be done in batches, then latency of hybrid system would go further down.

For SetFit models, we calculate OOS AUCROC by considering max predicted score amongst all labels. For black box LLMs, we calculate OOS AUCROC by considering score as 1 if LLM predicts an in-scope label, 0 otherwise.

\begin{table}[]
\centering
\resizebox{\columnwidth}{!}{%
\begin{tabular}{cc}
\hline
\textbf{Hyperparameter Name} & \textbf{Range of Values}    \\ \hline
body\_learning\_rate         & From 5e-6 till 5e-5         \\
head\_learning\_rate         & From 1e-3 till 1e-2         \\
num\_epochs                  & From 3 till 10              \\
batch\_size                  & Amongst {[}8, 16, 32, 64{]} \\
n\_trials                    & 10                          \\ \hline
\end{tabular}%
}
\caption{Hyperparameter search space for SetFit model training}
\label{tab:setfit_hyperparam}
\end{table}

\begin{table}[]
\centering
\resizebox{\columnwidth}{!}{%
\begin{tabular}{cc}
\hline
\textbf{Hyperparameter Name} & \textbf{Range of Values}     \\ \hline
k (no. of ICL examples)      & {[}0, 1, 5, 10, 20{]}        \\
t (retriever threshold)      & {[}0.00001, 0.3, 0.5, 0.7{]} \\ \hline
\end{tabular}%
}
\caption{Hyperparameter search space for choosing ICL examples for LLM based intent detection}
\label{tab:llm_hyperparam}
\end{table}

\begin{table*}[]
\centering
\resizebox{\textwidth}{!}{%
\begin{tabular}{ccc}
\hline
{\color[HTML]{000000} \textbf{Level 1 class}} & {\color[HTML]{000000} \textbf{Level 2 class}}                          & {\color[HTML]{000000} \textbf{Example Utterance}}                                      \\ \hline
{\color[HTML]{000000} Product Recommendation} & {\color[HTML]{000000} Static Product Attribute based}                  & {\color[HTML]{000000} show laptop with 8gb RAM}                                        \\
{\color[HTML]{000000} Product Recommendation} & {\color[HTML]{000000} Similarity/Comparison with other products based} & {\color[HTML]{000000} show laptop comparable to the Dell XPS 13}                       \\
{\color[HTML]{000000} Product Recommendation} & {\color[HTML]{000000} Compatibiliy with other products based}          & {\color[HTML]{000000} show laptop bags compatible with Dell XPS 15}                    \\
{\color[HTML]{000000} Product Recommendation} & {\color[HTML]{000000} Offers based}                                    & {\color[HTML]{000000} show laptop with HDFC bank EMI offers}                           \\
{\color[HTML]{000000} Product Recommendation} & {\color[HTML]{000000} Customer Reviews/Ratings based}                  & {\color[HTML]{000000} show laptops whose battery life is highly praised by users}      \\
{\color[HTML]{000000} Product Recommendation} & {\color[HTML]{000000} Budget based}                                    & {\color[HTML]{000000} show laptops under 50k}                                          \\
{\color[HTML]{000000} Product Recommendation} & {\color[HTML]{000000} Purpose/Usecase based}                           & {\color[HTML]{000000} show laptops suitable for graphic design work}                   \\
{\color[HTML]{000000} Product Recommendation} & {\color[HTML]{000000} Warranty/Return policy based}                    & {\color[HTML]{000000} show laptops with hassle-free return options}                    \\
{\color[HTML]{000000} Product Recommendation} & {\color[HTML]{000000} Delivery ETA based}                              & {\color[HTML]{000000} show laptops that can be delivered within the next week}         \\
{\color[HTML]{000000} Product Recommendation} & {\color[HTML]{000000} Past sales based}                                & {\color[HTML]{000000} show the most popular laptop models recently}                    \\
{\color[HTML]{000000} Product Evaluation}     & {\color[HTML]{000000} Static Product Attribute based}                  & {\color[HTML]{000000} does this laptop have 8gb RAM}                                   \\
{\color[HTML]{000000} Product Evaluation}     & {\color[HTML]{000000} Similarity/Comparison with other products based} & {\color[HTML]{000000} is this laptop comparable to the Dell XPS 13}                    \\
{\color[HTML]{000000} Product Evaluation}     & {\color[HTML]{000000} Compatibiliy with other products based}          & {\color[HTML]{000000} are these laptop bags compatible with Dell XPS 15}               \\
{\color[HTML]{000000} Product Evaluation}     & {\color[HTML]{000000} Offers based}                                    & {\color[HTML]{000000} does this laptop have HDFC bank EMI offers}                      \\
{\color[HTML]{000000} Product Evaluation}     & {\color[HTML]{000000} Customer Reviews/Ratings based}                  & {\color[HTML]{000000} are these laptops whose battery life is highly praised by users} \\
{\color[HTML]{000000} Product Evaluation}     & {\color[HTML]{000000} Budget based}                                    & {\color[HTML]{000000} are these laptops under 50k}                                     \\
{\color[HTML]{000000} Product Evaluation}     & {\color[HTML]{000000} Purpose/Usecase based}                           & {\color[HTML]{000000} are these laptops suitable for graphic design work}              \\
{\color[HTML]{000000} Product Evaluation}     & {\color[HTML]{000000} Warranty/Return policy based}                    & {\color[HTML]{000000} do these laptops have hassle-free return options}                \\
{\color[HTML]{000000} Product Evaluation}     & {\color[HTML]{000000} Delivery ETA based}                              & {\color[HTML]{000000} can these laptops be delivered within the next week}             \\
{\color[HTML]{000000} Product Evaluation}     & {\color[HTML]{000000} Past sales based}                                & {\color[HTML]{000000} are these the most popular laptop models recently}               \\ \hline
\end{tabular}%
}
\caption{Example utterance for each leaf intent from controlled experiment dataset used to understand behavior of LLM based intent detection.}
\label{tab:controlled_exp_dataset}
\end{table*}

\subsection{Controlled Experiment}
\label{appendix:controlled_exp}

\textbf{Setup.} For our controlled experiment dataset, we hand-curate 10 utterances per leaf intent, random 5 of which we use in train and other 5 we use in test for every run. We also use three paraphrases (pre-curated) of each test utterance in our test set for every run to test generalization across utterance variants. For controlled experiment, we train all SetFit models with batch size of 16 and 5 epochs. For ICL examples selection with LLMs, we use max 5 ICL examples with retriever threshold of 1e-5. Since we execute every experiment 10 times with randomly created dataset, we are unable to experiment with other hyperparameters due to compute costs. Since we do controlled experiments to develop better understanding of LLM behavior, keeping these hyper-parameters fixed is okay. 

\begin{figure*}
\centering
\begin{subfigure}{.5\linewidth}
  \centering
  \includegraphics[width=0.9\linewidth,height=4.5cm]{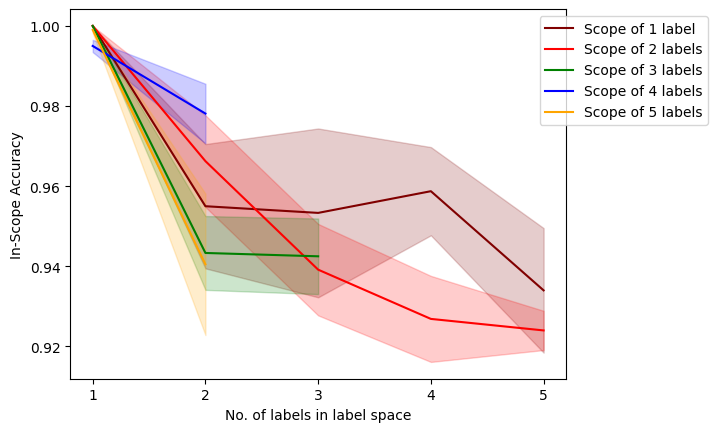}
  \caption{Claude v3 Sonnet}
  \label{fig:controlled_in_scope_acc_claude_v3_sonnet}
\end{subfigure}%
\begin{subfigure}{.5\linewidth}
  \centering
  \includegraphics[width=0.9\linewidth,height=4.5cm]{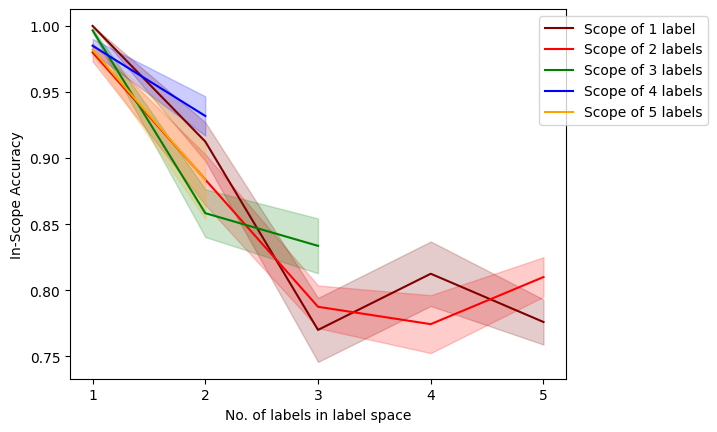}
  \caption{Claude v3 Haiku}
  \label{fig:controlled_in_scope_acc_claude_v3_haiku}
\end{subfigure}
\begin{subfigure}{.5\linewidth}
  \centering
  \includegraphics[width=0.9\linewidth,height=4.5cm]{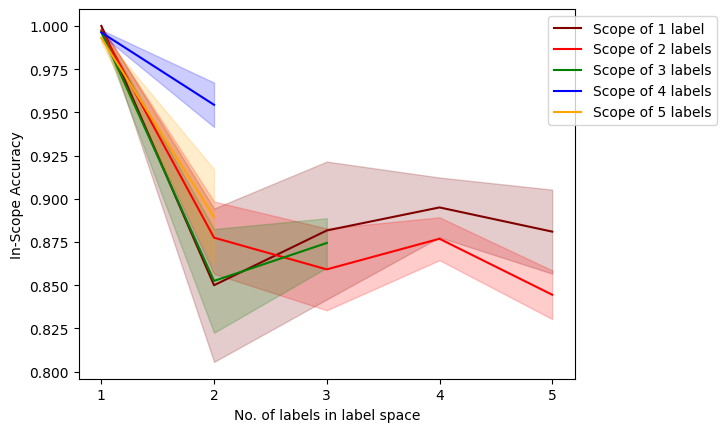}
  \caption{Mistral 7B}
  \label{fig:controlled_in_scope_acc_mistral_7b}
\end{subfigure}%
\begin{subfigure}{.5\linewidth}
  \centering
  \includegraphics[width=0.9\linewidth,height=4.5cm]{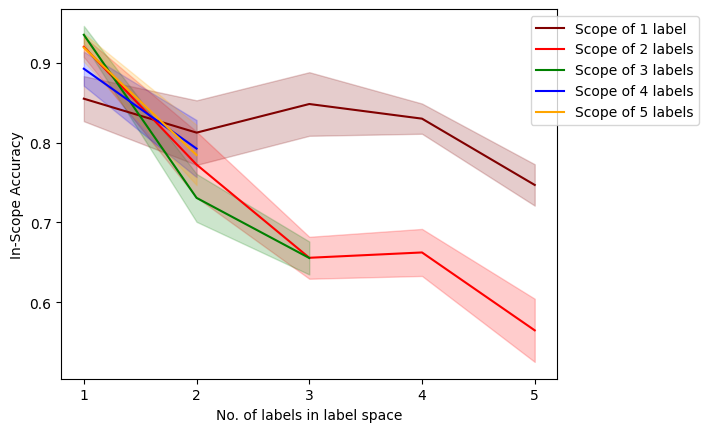}
  \caption{SetFit + Neg Aug}
  \label{fig:controlled_in_scope_acc_baseline}
\end{subfigure}
\caption{Change in In-Scope accuracy with number of labels in label space and scope of labels.}
\label{fig:controlled_in_scope_acc}
\end{figure*}

\begin{figure*}
\centering
\begin{subfigure}{.5\linewidth}
  \centering
  \includegraphics[width=0.9\linewidth,height=4.5cm]{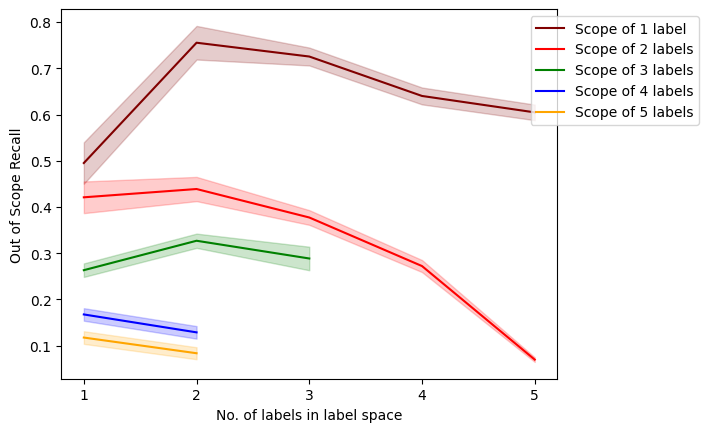}
  \caption{Claude v3 Sonnet}
  \label{fig:controlled_oos_recall_claude_v3_sonnet}
\end{subfigure}%
\begin{subfigure}{.5\linewidth}
  \centering
  \includegraphics[width=0.9\linewidth,height=4.5cm]{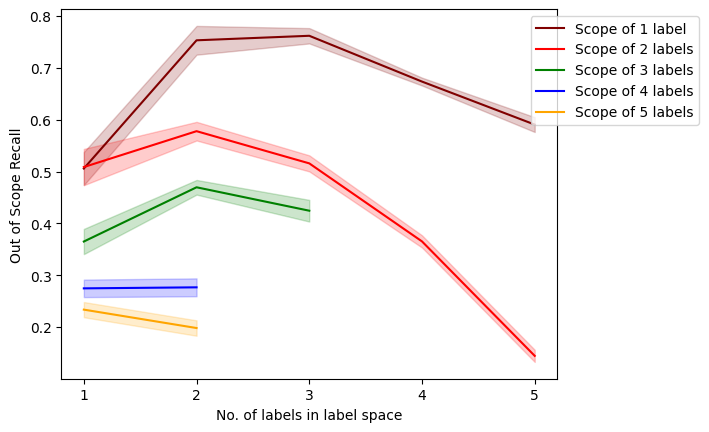}
  \caption{Claude v3 Haiku}
  \label{fig:controlled_oos_recall_claude_v3_haiku}
\end{subfigure}
\begin{subfigure}{.5\linewidth}
  \centering
  \includegraphics[width=0.9\linewidth,height=4.5cm]{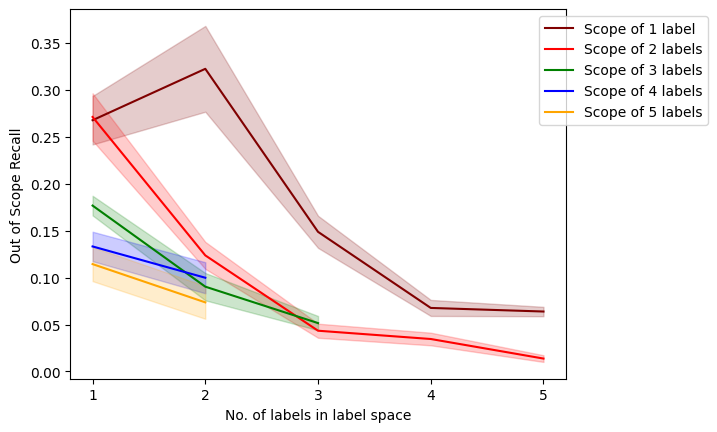}
  \caption{Mistral 7B}
  \label{fig:controlled_oos_recall_mistral_7b}
\end{subfigure}%
\begin{subfigure}{.5\linewidth}
  \centering
  \includegraphics[width=0.9\linewidth,height=4.5cm]{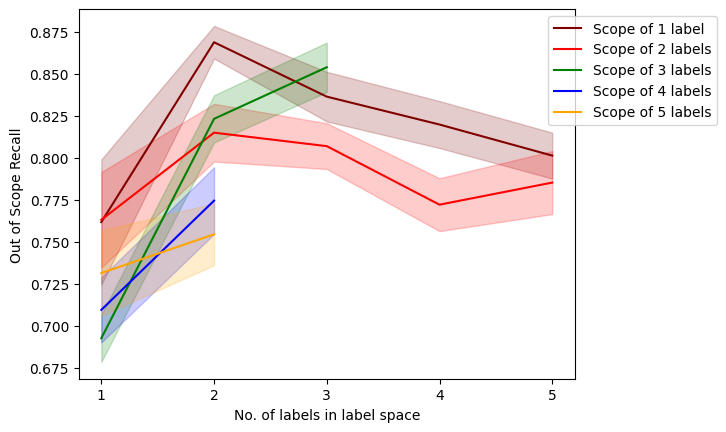}
  \caption{SetFit + Neg Aug}
  \label{fig:controlled_oos_recall_baseline}
\end{subfigure}
\caption{Change in OOS Recall with number of labels in label space and scope of labels.}
\label{fig:controlled_oos_recall}
\end{figure*}

\noindent\textbf{Results.} Table \ref{tab:controlled_exp_dataset} shows example queries from each intent from controlled experiment dataset.
From controlled experiments, Fig \ref{fig:controlled_in_scope_acc} and Fig \ref{fig:controlled_oos_recall} show change in In-Scope accuracy and OOS Recall with number of labels in label space and scope of labels, respectively.

\end{document}